\documentclass{article}
\usepackage{spconf,amsmath,graphicx}
\usepackage{booktabs,array,multirow}
\usepackage[table]{xcolor}
\definecolor{famshade}{gray}{0.90}

\title{Do Speech Representations Preserve Regional Accent Across Read and Spontaneous Speech?}

\name{
\shortstack{
Paula A. Perez-Toro$^{1}$ \qquad Tomas Arias-Vergara$^{1}$ \qquad Annette Schwarz$^{1}$ \qquad Abner Hernandez$^{1}$\\
Andreas Horr$^{2}$ \qquad Cornelia Kristen$^{3}$ \qquad Andreas Maier$^{1}$
}
\thanks{Corresponding author: paula.andrea.perez@fau.de}
}

\address{\fontsize{9.5}{11}\selectfont
$^{1}$Pattern Recognition Lab, Friedrich-Alexander-Universit\"at Erlangen-N\"urnberg, Erlangen, Germany. \fontsize{9.5}{11}\selectfont
$^{2}$Leibniz Institute for \\ \fontsize{9.5}{11}\selectfont Educational Trajectories (LIfBi), Bamberg, Germany. 
$^{3}$Otto-Friedrich-Universit\"at Bamberg, Bamberg, Germany.
}

\begin{document}
\sloppy
\maketitle
\begin{abstract}
Regional accent cues can be captured under matched conditions, but it remains unclear whether they persist between read and spontaneous speech. We study RVG1, with 500 German speakers from nine regions, comparing ten speech representations on regional classification and continuous geolocation under matched conditions and speaker-independent read--spontaneous transfer. Whisper performs best under matched conditions, reaching 0.489 nine-way UAR and 148~km median geolocation error, but drops to 0.11/0.18 UAR across transfer directions and 363~km geolocation error. Self-supervised models show a similar degradation, whereas speaker embeddings are less discriminative in-domain but more robust under transfer. This contrast is consistent across classification and geolocation. Across representations, robustness is associated with how little a representation shifts between styles (style-invariance), for which crossstyle speaker retrieval is an interpretable proxy. Age, sex, sentence-overlap, and duration controls do not account for the gap, although channel characteristics contribute. These results show that strong matched-condition performance does not indicate robust regional information.
\end{abstract}
\begin{keywords}
German Regional Accent, Speech Representations, Geolocation

\end{keywords}
\section{Introduction}
\label{sec:intro}
\begin{figure*}[t]
\centering
\includegraphics[width=0.88\textwidth]{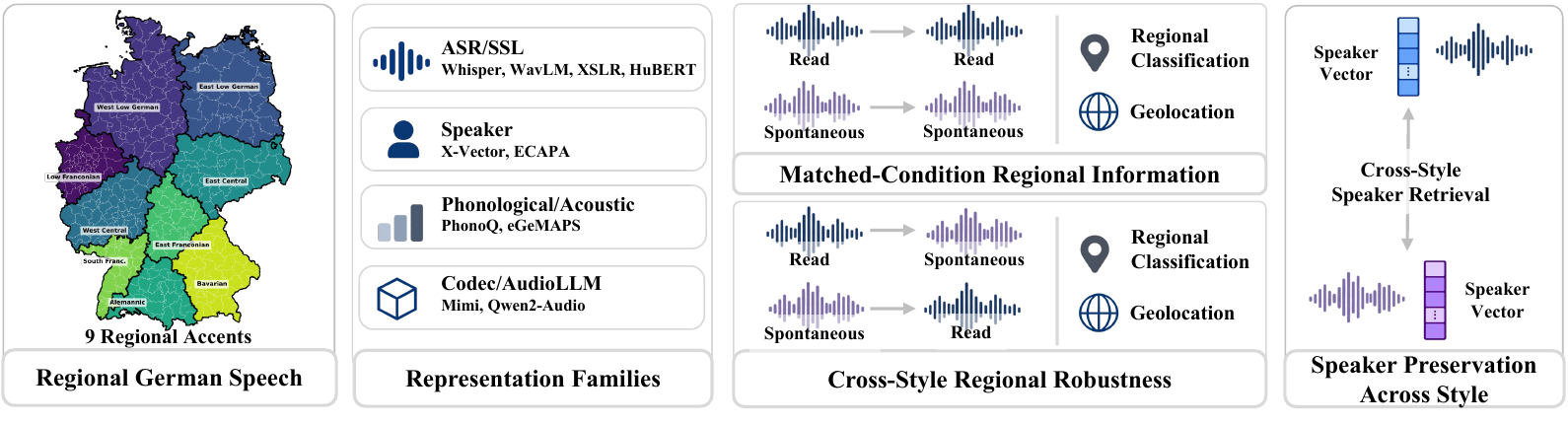}
\caption{Overview of the study. Ten speech representations are evaluated on German regional classification and geolocation under matched conditions and speaker-independent read–spontaneous transfer.} 
\label{fig:overview}
\end{figure*}

Regional variation affects pronunciation, prosody, and lexical choice, making geographic origin detectable and introducing variability in speech-processing systems. It has long been studied in dialectometry~\cite{wieling2015} and is relevant to forensic speaker profiling~\cite{jessen2008} and privacy-preserving speech processing~\cite{nautsch2019}. Regional Variants of German 1 (RVG1) contains read and spontaneous speech from approx. 500 speakers across the German-speaking area (including Austrian, Swiss, and northern-Italian speakers)~\cite{burger1998rvg1}. Earlier RVG1/j studies using GMM--UBM systems reached class-balanced accuracies of $35.3\%$ for nine regions and $56.0\%$ for three broad regions~\cite{dobbriner2019slate}, while phonotactic experiments reported a maximum accuracy of $19.2\%$ for nine spontaneous-speech dialects~\cite{stadtschnitzer2018}.

Recent work uses pretrained speech representations for accent and dialect recognition. Self-Supervised Learning (SSL) models such as wav2vec~2.0, HuBERT, and WavLM provide general-purpose speech representations~\cite{baevski2020wav2vec2,hsu2021hubert,chen2022wavlm}, while XLSR and Whisper provide multilingual and large-scale ASR representations~\cite{conneau2021xlsr,radford2023whisper}. Multilingual SSL models have been adapted to language and dialect identification~\cite{shaik2024sapt} and continuous and discrete SSLs explored for accent recognition~\cite{wang2025accent}. Voxlect further evaluates speech foundation models for dialect and regional-language classification across languages, including German~\cite{feng2026voxlect}.
It remains unclear whether regional information remains stable when speech changes from read to spontaneous. Linke et al. showed that speaking style can strongly structure SSL representations, with style similarity outweighing regional-variety similarity in some cases~\cite{linke23}. Geographic variation can also be modeled continuously through speech geolocation rather than predefined classes~\cite{foley2024geolocation,khadraoui2026}. 

However, strong performance when training and testing on the same speech type does not establish robustness under read--spontaneous transfer. Our contributions are:
(1) We introduce a controlled framework to evaluate whether regional information transfers between read and spontaneous speech, with speaker-independent evaluation under matched and mismatched conditions.
(2) We compare ten representations from seven model families for regional classification and continuous geolocation, complemented by analyses of regional granularity and potential confounds.
(3) We investigate style-invariance (how little a representation shifts between read and spontaneous speech) as a factor underlying robustness, relating cross-condition speaker retrieval and representation shift to classification and geolocation.

Throughout the paper, we use \emph{regional accent} for geographically conditioned variation in the speech signal.

\section{Data and Methods}
\label{sec:methods}
\vspace{-5pt}

\textbf{Corpus and labels:} RVG1-v3 provides 500 region-labelled German speakers with read speech, 498 of whom also have spontaneous recordings and 432 geocoded hometown coordinates; accordingly, classification uses 500 read and 498 spontaneous speakers, read--spontaneous analyses 498, and geolocation 432. The nine classes are imbalanced (speaker counts A--I: 44/103/31/73/53/63/20/13/100), motivating unweighted average recall (UAR) as the primary classification metric. We use the corpus taxonomy as the finest target (A:~Low Franconian, B:~West Low German, C:~East Low German, D:~West Central German, E:~East Central German, F:~Alemannic, G:~East Franconian, H:~South Franconian, I:~Bavarian) and additionally consider six analysis-specific geographic classes ($\{A,B\}$, $\{C\}$, $\{D\}$, $\{E\}$, $\{F,G,H\}$, $\{I\}$), a 3-way partition $\{A,B,C\}$, $\{D,E\}$, and $\{F,G,H,I\}$, and a 2-way Low- versus High-German partition following the Benrath-line division~\cite{wiesinger1983}. Region labels are treated as evaluation categories rather than sharp linguistic dialect boundaries. Read speech consists mainly of digit, number, and command strings plus phonetically balanced sentences, whereas spontaneous speech is a prompted monologue imagining an address to someone from the speaker's home region; read--spontaneous transfer therefore also crosses a register boundary.


\noindent\textbf{Representations:} We compare ten frozen speaker-level representations spanning seven families, evaluated for classification and geolocation, in-domain and cross-style: Whisper-large-v3 encoder features~\cite{radford2023whisper} (ASR); WavLM-large~\cite{chen2022wavlm}, wav2vec~2.0-XLSR~\cite{baevski2020wav2vec2,conneau2021xlsr}, and HuBERT-large~\cite{hsu2021hubert} (SSL); speaker embeddings, x-vector~\cite{snyder2018} and ECAPA-TDNN~\cite{desplanques2020ecapa} (SpeechBrain \texttt{spkrec-xvect-voxceleb}, $512$-d, and \texttt{spkrec-ecapa-voxceleb}, $192$-d, both VoxCeleb-trained); a 97-dimensional PhonoQ phonological descriptor~\cite{ariasvergara2022}; eGeMAPS acoustic features~\cite{eyben2016egemaps}; the $512$-dimensional continuous pre-quantization encoder latent of the Mimi audio codec~\cite{defossez2024moshi} (\texttt{kyutai/mimi}, $12.5$~Hz frames); and the audio-encoder hidden states ($1280$-d) of the Qwen2-Audio-7B-Instruct audio LLM~\cite{chu2024qwen2audio}. PhonoQ computes phoneme-posterior probabilities for 18 phonological classes and summarizes by maximum-posterior and log-likelihood-ratio statistics, average phone duration and its variability, and raw/normalized pairwise-variability (rPVI/nPVI) rhythm measures, aggregated to one vector per speaker; it is thus an interpretable phonological descriptor rather than a pretrained encoder. 
We mean-pool encoder activations into one vector per speaker and use fixed layers selected by preliminary nine-way probing: Whisper L23, WavLM L24, wav2vec2-XLSR L10, and HuBERT L13. 
A fully nested layer-selection check using an untuned logistic-regression probe gave similar 9-way UAR to the fixed-layer choice (mean 0.429 vs.\ 0.428 across encoders; Whisper 0.512 vs.\ 0.510), indicating that the fixed selection does not benefit from leakage. 
Thus, we use the fixed-layer protocol throughout; these probe scores are not directly comparable to the tuned MLP results in Tab.~\ref{tab:main}.

\noindent \textbf{Evaluation:} We use five speaker-independent $70{:}30$ train/test splits, fitting preprocessing on training only. 
Classifier parameters are tuned by 3-fold cross-validation on the train split, with MLP hidden sizes of 128/256. 9-way representation differences are assessed with paired t-tests across the five matched splits and Holm--Bonferroni correction; because the five $70{:}30$ splits overlap in speakers, these tests are treated as descriptive supporting evidence rather than exact inference.

\begin{table*}[htpb]
\centering
\caption{Regional classification UAR\,/\,ACC values on read speech. 
Prior published RVG1 results are listed for context only (different partitions/preprocessing; not a controlled comparison).}
\label{tab:main}
\renewcommand{\arraystretch}{0.95}
\setlength{\tabcolsep}{26pt}
\resizebox{\linewidth}{!}{
\begin{tabular}{llcccc}
\toprule
Family & Representation & 9-way & 6-way & 3-way & 2-way \\
\midrule
\rowcolor{famshade} ASR & \textbf{Whisper} & \textbf{0.489 / 0.568} & \textbf{0.627 / 0.691} & \textbf{0.767 / 0.788} & \textbf{0.835 / 0.841} \\
SSL & WavLM & 0.397 / 0.480 & 0.504 / 0.565 & 0.702 / 0.720 & 0.775 / 0.792 \\
 & wav2vec2-XLSR & 0.391 / 0.488 & 0.486 / 0.548 & 0.698 / 0.719 & 0.800 / 0.809 \\
 & HuBERT & 0.385 / 0.469 & 0.492 / 0.561 & 0.664 / 0.688 & 0.762 / 0.783 \\
\rowcolor{famshade} Audio-LLM & Qwen2-Audio & 0.323 / 0.381 & 0.419 / 0.465 & 0.616 / 0.647 & 0.798 / 0.780 \\
Speaker & x-vector & 0.270 / 0.352 & 0.390 / 0.455 & 0.619 / 0.640 & 0.741 / 0.749 \\
 & ECAPA & 0.243 / 0.296 & 0.350 / 0.413 & 0.606 / 0.636 & 0.719 / 0.708 \\
\rowcolor{famshade} Phonological & PhonoQ & 0.268 / 0.335 & 0.379 / 0.416 & 0.549 / 0.565 & 0.695 / 0.704 \\
Acoustic & eGeMAPS & 0.253 / 0.284 & 0.325 / 0.380 & 0.480 / 0.501 & 0.653 / 0.665 \\
\rowcolor{famshade} Codec & Mimi & 0.220 / 0.264 & 0.349 / 0.353 & 0.431 / 0.445 & 0.649 / 0.651 \\
\midrule
\multicolumn{6}{l}{\emph{Prior published RVG1 (context only; ACC)}} \\
 & GMM--UBM~\cite{dobbriner2019slate} & 0.14--0.35 & -- & 0.38--0.56 & -- \\
\bottomrule
\end{tabular}}
\end{table*}

\noindent\textbf{Controls and auxiliary tasks:} We assess demographic confounding using age and sex alone, by adding them to the speech features, and by residualizing them from the representations. Microphone robustness is evaluated over the $4\times4$ train/test matrix. For geolocation, all representations use five speaker-disjoint splits with training-only preprocessing; high-dimensional features are reduced to 256 principal components
(via PCA, fit on training speakers) when needed. Coordinates are predicted with a two-layer MLP trained using great-circle loss and evaluated by median great-circle distance. Read--spontaneous transfer uses held-out speakers, with sentence-independent and duration-matched controls to test prompt overlap and speech-duration effects.

\noindent\textbf{Cross-style speaker retrieval:} 
To measure preservation of speaker identity across style (read vs. spontaneous), we mean-pool representations within utterances and average them by speaker and style, yielding one $\ell_2$-normalized vector per speaker/style. 
Each spontaneous-speech vector queries read-speech vectors by cosine similarity, with rank-1 accuracy given by the fraction whose nearest neighbor is the same speaker (498 queries; chance = 1/498 $\approx$ 0.002).
No model is fit; speaker labels only define matches and score retrieval.


\section{Results}
\label{sec:results}
\vspace{-5pt}

\subsection{Matched-condition regional discrimination}

\indent \textbf{Representation and granularity:} Whisper ranks first at every granularity and the family-level ordering is preserved, although ranks below the top change across granularities (Tab.~\ref{tab:main}). Whisper reaches a 9-way UAR of $0.489$ and a 3-way UAR of $0.767$. Its 9-way advantage is consistent across all five matched splits and all nine other representations (paired differences favoring Whisper on every split; treated as descriptive, see Sec.~\ref{sec:methods}). Mean gains range from about $0.09$ UAR over the strongest SSL representation to $0.27$ over Mimi. Performance increases monotonically for all ten representations as the label space is coarsened. An untuned logistic-regression probe yields a very similar ordering, suggesting that the ranking is not driven by classifier tuning. Matched spontaneous-speech performance follows a similar ranking across granularities (Whisper $0.431$, $0.572$, $0.756$, $0.832$ for 9-, 6-, 3-, and 2-way, respectively), indicating that the degradation in Fig.~\ref{fig:transfer}b reflects read--spontaneous transfer rather than an inherent difficulty of spontaneous speech.

\begin{figure}[!hbtp]
\centering
\includegraphics[width=\linewidth]{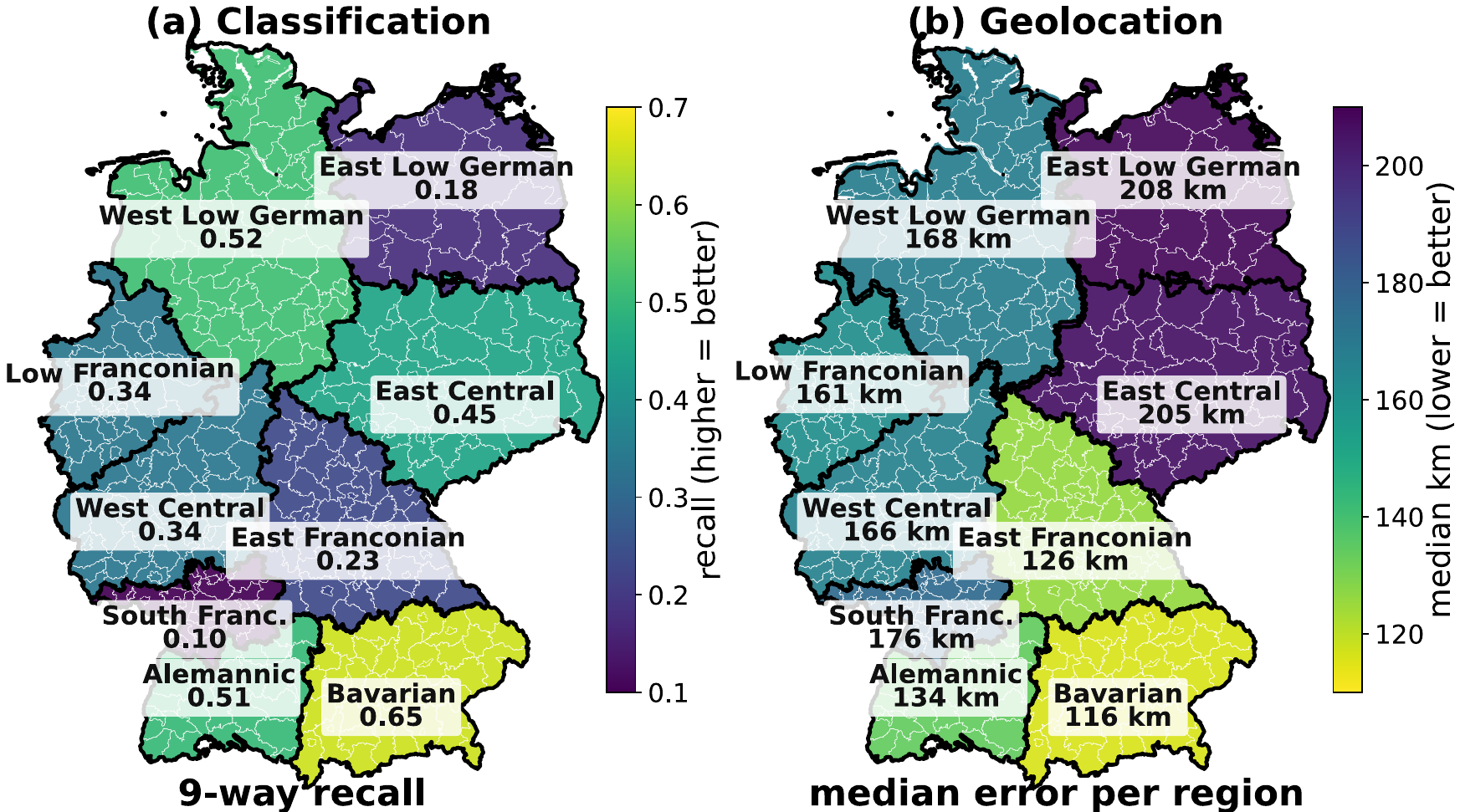}
\caption{Regional heterogeneity of the Whisper representation. Recall ranges from $0.10$ (South Franconian) to $0.65$ (Bavarian) and localization error from $116$ to $208$~km.
}
\label{fig:maps}
\vspace{-13pt}
\end{figure}

\noindent\textbf{In-domain geolocation and regional structure:} Whisper is the strongest matched-condition geolocation representation (Fig.~\ref{fig:transfer}d), reaching $148\pm16$~km ($\pm$ denotes the standard
deviation across the five splits) with a two-layer MLP trained using great-circle loss, compared with $175$~km for direct coordinate regression and $242$~km for the mean-location baseline. An auxiliary regional-classification head provides no further gain ($148$~km; $p=0.42$). Region-level 9-way recall correlates strongly with class size (Spearman $\rho=0.87$, $p=0.003$), ranging from $0.10$ for South Franconian to $0.65$ for Bavarian, while median geolocation error ranges from $116$ to $208$~km (Fig.~\ref{fig:maps}). Leave-one-region-out errors show geographic structure, but because recall tracks class size, the maps are descriptive rather than evidence of differences in dialect difficulty.

\subsection{Robustness to speaking-style shift}


\begin{figure*}[htpb]
\centering
\includegraphics[width=0.805\textwidth]{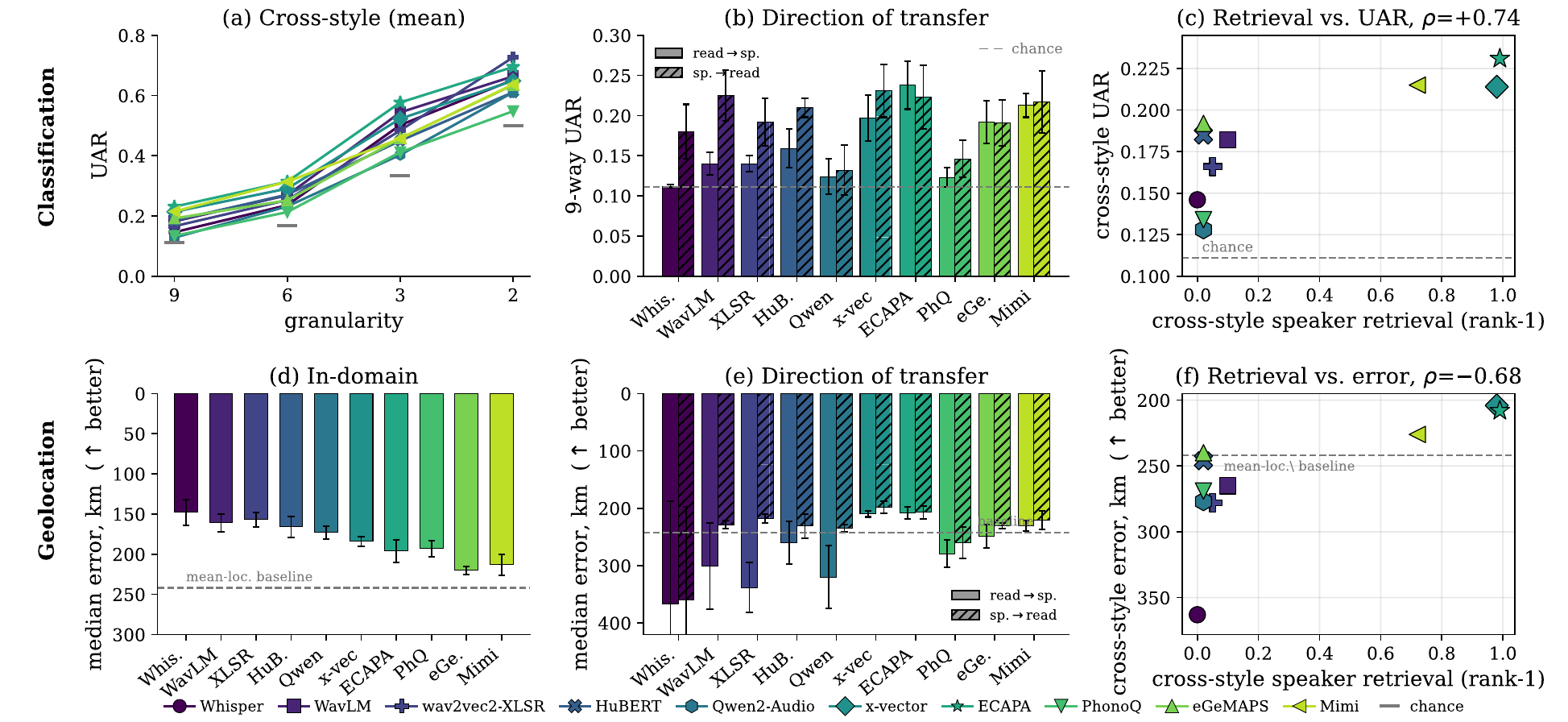}
\caption{Read--spontaneous transfer for classification (top) and geolocation (bottom).
\emph{(a)}~Cross-condition UAR by granularity, with a 9-way ranking reversal.
\emph{(b)}~9-way transfer by direction; content encoders degrade most.
\emph{(c)}~Speaker retrieval vs.\ 9-way UAR ($\rho=+0.74$).
\emph{(d)}~In-domain geolocation error (Whisper: $148$~km).
\emph{(e)}~Geolocation transfer by direction; speaker embeddings and Mimi remain near baseline.
\emph{(f)}~Speaker retrieval vs.\ transfer error ($\rho=-0.68$; inverted $y$-axis).}
\label{fig:transfer}
\vspace{-8pt}
\end{figure*}

\noindent\textbf{Read--spontaneous degradation:} Transfer performance varies strongly by representation family on both tasks (Fig.~\ref{fig:transfer}). For Whisper, 9-way UAR falls to $0.11$ for read$\rightarrow$spontaneous and $0.18$ for spontaneous$\rightarrow$read (Fig.~\ref{fig:transfer}b), while geolocation error rises from $148$ to $363$~km (Fig.~\ref{fig:transfer}d--e), exceeding the $242$~km mean-location baseline. SSL encoders show similar degradation (wav2vec2-XLSR $0.14$/$0.19$ UAR), whereas speaker embeddings and the codec retain more regional information (ECAPA $0.25\rightarrow0.23$ UAR, $196\rightarrow208$~km). Sentence-disjoint evaluation reduces Whisper 9-way UAR from about $0.48$ to $0.40$ (3-way $0.77$ to $0.73$), while duration matching leaves read UAR unchanged ($0.486\rightarrow0.488$) and transfer near $0.11$. Thus, prompt overlap and speech amount alone do not explain the gap.

\noindent\textbf{Style invariance is associated with robustness.} Rank-1 speaker retrieval between read and spontaneous speech is strongly associated with transfer performance across the ten representations: classification (Spearman $\rho=+0.74$, $p=0.014$; Fig.~\ref{fig:transfer}c) and geolocation ($\rho=-0.68$, $p=0.030$; Fig.~\ref{fig:transfer}f). Speaker embeddings reach $\approx\!0.99$ retrieval and the codec $0.72$, whereas content-oriented representations remain at $\leq0.10$, suggesting that regional information is better retained when speaker-related information is preserved across conditions. For geolocation, lower in-domain error is also associated with higher transfer error ($\rho=-0.71$, $p=0.022$), robust to leave-one-representation-out analysis and removal of both speaker embeddings. This trade-off is weaker for classification ($\rho=-0.37$, n.s.).
With only ten representations from partly related families, we treat all four correlations as exploratory: jointly, they do not survive Holm (smallest $p = 0.014 > 0.05/4$). The two central associations are stable under leave-one-representation-out ($\rho \in [0.64, 0.81]$ for classification; $\rho \in [-0.85, -0.64]$ for geolocation) but weaken to non-significance when both speaker embeddings are removed.
Moreover, auxiliary multi-task objectives did not significantly improve geolocation or transfer.

\subsection{Controls, confounds, and further analysis}
Age and sex alone yield 9-way UAR $0.141$, close to chance, and adding them to Whisper does not improve regional UAR ($0.489\rightarrow0.489$). Residualizing age and sex from the features changes regional UAR by less than $0.02$, and regional performance remains well above chance within each sex (male $0.49$, female $0.43$). At the same time, Whisper strongly encodes demographic information, reaching $0.99$ balanced accuracy for sex and $4.4$-year age mean absolute error (MAE; against about 7 years for a constant predictor, since 66\% of speakers are aged 21–30). The controls therefore support a limited conclusion: age and sex are encoded, but neither variable alone accounts for the observed regional discrimination. 
Furthermore, averaged across mismatched microphone pairs, UAR remains above chance at $0.27$ for 9-way and $0.60$ for 3-way, but both are lower than the corresponding matched-microphone values of $0.36$ and $0.68$. Channel characteristics therefore contribute to performance, even though regional information remains detectable under microphone mismatch.


\section{Discussion \& Conclusion}
\vspace{-5pt}
The results distinguish matched-condition \emph{discriminability} from robustness across read and spontaneous speech. Whisper performs best for regional classification and geolocation under matched conditions, and coarse 3-way structure is easier to recover than fine nine-region distinctions. Age and sex do not explain this performance, and regional information remains detectable under microphone mismatch, although channel effects remain.
The central finding is that read--spontaneous robustness is more closely associated with style-invariance (how little a representation shifts across styles, proxied by cross-style speaker retrieval) than with matched-condition strength. Whisper, the SSL models, and Qwen2-Audio perform strongly in-domain but degrade markedly under transfer, whereas speaker embeddings and, to a lesser extent, Mimi remain more stable. For geolocation, the strongest in-domain representations are also among the least robust. This complements prior work on pretrained representations for accent and dialect recognition~\cite{shaik2024sapt,wang2025accent} by linking robustness to style-invariance of the representation across speech conditions. These findings suggest that accent systems should be evaluated explicitly in both read and spontaneous speech, rather than relying solely on the performance of the matched-conditions. They also motivate future work on whether speaker-oriented representations retain geographic information relevant to anonymization. The study is limited to one imbalanced German corpus, and a broader validation will require additional datasets.

\section{Compliance with Ethical Standards}
This study uses the pre-existing corpus; no new participants were recruited or data collected. The RVG1 corpus was used under the research-use terms of the Bavarian Archive for Speech Signals.


\bibliographystyle{IEEEbib}
\bibliography{references}

\end{document}